\documentclass[10pt,twocolumn,letterpaper]{article}

\usepackage{iccv}
\usepackage{times}
\usepackage{epsfig}
\usepackage{graphicx}
\usepackage{amsmath}
\usepackage{amssymb}

\usepackage{multirow}
\usepackage[usestackEOL]{stackengine}

\usepackage{caption}
\newcommand{\squishlist}{
 \begin{list}{$\bullet$}
  { \setlength{\itemsep}{0pt}
     \setlength{\parsep}{2pt}
     \setlength{\topsep}{2pt}
     \setlength{\partopsep}{0pt}
     \setlength{\leftmargin}{1.5em}
     \setlength{\labelwidth}{1em}
     \setlength{\labelsep}{0.5em} 
  } 
}
\newcommand{\squishend}{\end{list}}

\usepackage[pagebackref=true,breaklinks=true,letterpaper=true,colorlinks,bookmarks=false]{hyperref}

\iccvfinalcopy 

\def\iccvPaperID{195} 

\ificcvfinal\fi
\begin{document}

\title{Recurrent Models for Situation Recognition}

\author{Arun Mallya and Svetlana Lazebnik\\
University of Illinois at Urbana-Champaign\\
{\tt\small \{amallya2,slazebni\}@illinois.edu}
}


\maketitle

\begin{abstract}
This work proposes Recurrent Neural Network (RNN) models to predict structured \lq image situations\rq\ -- actions and noun entities fulfilling semantic roles related to the action. In contrast to prior work relying on Conditional Random Fields (CRFs), we use a specialized action prediction network followed by an RNN for noun prediction. Our system obtains state-of-the-art accuracy on the challenging recent imSitu dataset, beating CRF-based models, including ones trained with additional data. Further, we show that specialized features learned from situation prediction can be transferred to the task of image captioning to more accurately describe human-object interactions. 
\end{abstract}

\thispagestyle{empty}
\vspace{-4mm}
\section{Introduction}
\label{sec:Introduction}

Recognition of actions and human-object interactions in still images has been widely studied in computer vision. 
Early datasets and approaches focused on identifying a relatively small number of actions, such as 10 in PASCAL VOC~\cite{everingham2010pascal} and 40 in the Stanford Dataset~\cite{yao2011human}. Newer and larger datasets such as MPII Human Pose~\cite{pishchulin2014fine} have enlarged the number of action classes to around 400. 
The COCO-A~\cite{ronchi2015COCOA} and HICO~\cite{chao2015hico} datasets aim to recognize interactions between multiple humans, and humans and objects, expanding the scope of recognition to outputs such as \emph{human-riding-bicycle}, \emph{human-repairing-bicycle}, \emph{human-riding-horse}, \etc.

\begin{figure}[t!]
  \includegraphics[trim={0.1cm 6.9cm 14cm 0cm},clip,scale=0.77]{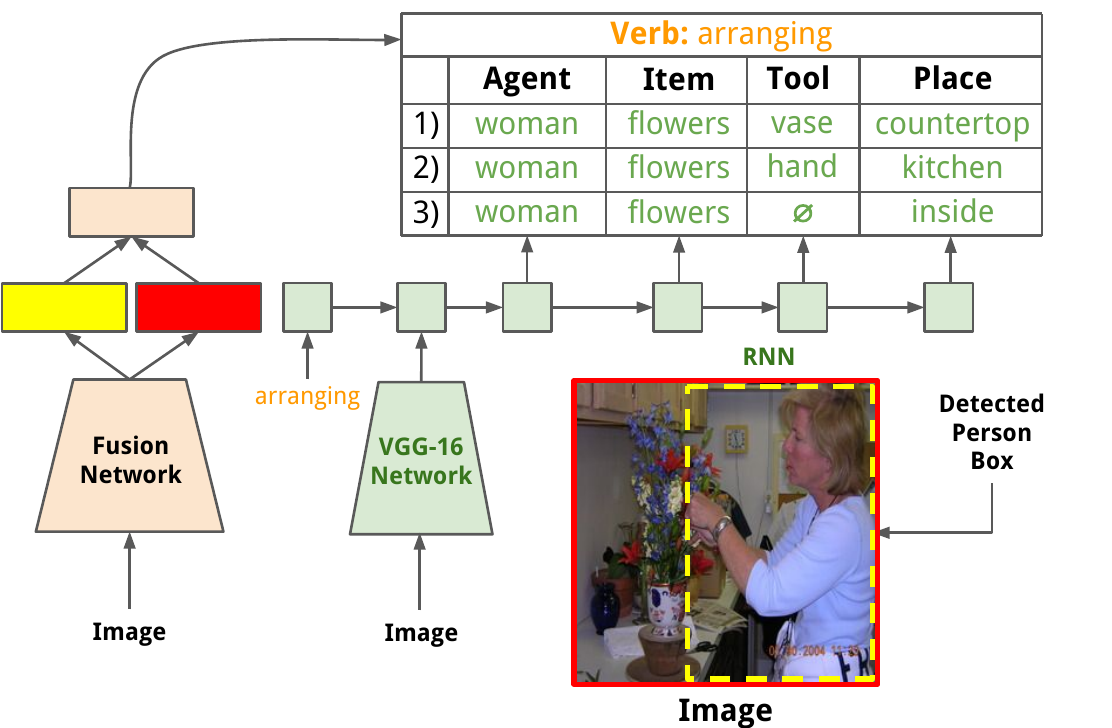}
  \caption{Each image in imSitu is labeled with an action verb (orange), and each verb is associated with a unique set of semantic roles (bold black) which are fulfilled by noun entities present in the image (green). Each image has multiple annotations to account for the intrinsic ambiguity of the task. Our approach first uses the fusion network of~\cite{mallya2016learning} to predict the action verb. Then it feeds the verb and a visual feature from a separate network into an RNN to predict the noun roles in a fixed sequence conditioned on the action.}
  \label{fig:overview}
\end{figure}

Of late, the focus has shifted to predicting even more structured outputs, tackling higher-level questions such as \emph{who} is doing \emph{what} and with \emph{which object}. The recently introduced imSitu Dataset~\cite{yatskar2016} generalizes the task of action recognition to \lq situation recognition\rq\ --- the recognition of all entities fulfilling semantic roles in an instance of an action performed by a human or non-human actor. Given a particular action, situations are represented by a set of relevant (semantic role: noun entity) pairs. An example image and associated situation from imSitu are shown in Fig.~\ref{fig:overview}, where \lq\lq a woman arranging flowers in a vase on the countertop\rq\rq\ is represented by \emph{Action}: arranging, \{(\emph{Agent}: woman), (\emph{Item}: flowers), (\emph{Tool}: vase), (\emph{Place}: countertop)\}. As another example, \lq\lq A horse rearing outside\rq\rq\ can be mapped to \emph{Action}: rearing, \{(\emph{Agent}: horse), (\emph{Place}: outside)\}. imSitu consists of 504 actions, 1,700 semantic roles, and 11,000 noun entities resulting in around 200,000 unique situations. 
Along with the dataset, Yatskar \etal~\cite{yatskar2016,yatskar_semsparsity} also introduced Conditional Random Field (CRF) models to predict situations given an image.  
In our work, we propose and train Recurrent Neural Networks (RNNs) to predict such situations and outperform the previously state of the art CRFs. 

Our use of RNNs for situation prediction is motivated by their popularity for tasks like image caption generation, where they have proven to be successful at capturing grammar and forming coherent sentences linking multiple concepts.  
The standard framework for caption generation involves feeding high-level features from a CNN, often trained for image classification on ImageNet~\cite{ILSVRC15}, into an RNN that proceeds to generate one word of the caption at a time~\cite{karpathy2015deep,vinyals2015show,vinyals2016show,donahue2014long,you2016image}. 
Situation recognition involves the prediction of a sequence of noun entities for a particular action, so it can be viewed as a more structured version of the captioning task with a grammar that is fixed given an action. 

Figure \ref{fig:overview} gives an overview of our best proposed system. First, we predict the action verb using the specialized action recognition architecture of~\cite{mallya2016learning}, which fuses features from a detected person box with a global representation of the image. Conditioned on the action, we treat the prediction of noun entities as a sequence generation problem and use an RNN. Details of our model, along with several baselines, will be given in Section \ref{sec:method}. Through extensive experiments (Section \ref{sec:results_imsitu}) we found that using separate networks for predicting the action verb and the noun entities produces higher accuracy than jointly training a visual representation for the two tasks. Finally, in Section~\ref{sec:coco} we explore how knowledge gained from situation prediction can obtain meaningful improvements for image captioning on the MSCOCO dataset~\cite{lin2014microsoft} through feature transfer.


\section{The Situation Prediction Task and Methods}
\label{sec:method}

\begin{figure*}[t!]
  \includegraphics[trim={0cm 10.5cm 2cm 0cm},clip,width=\textwidth]{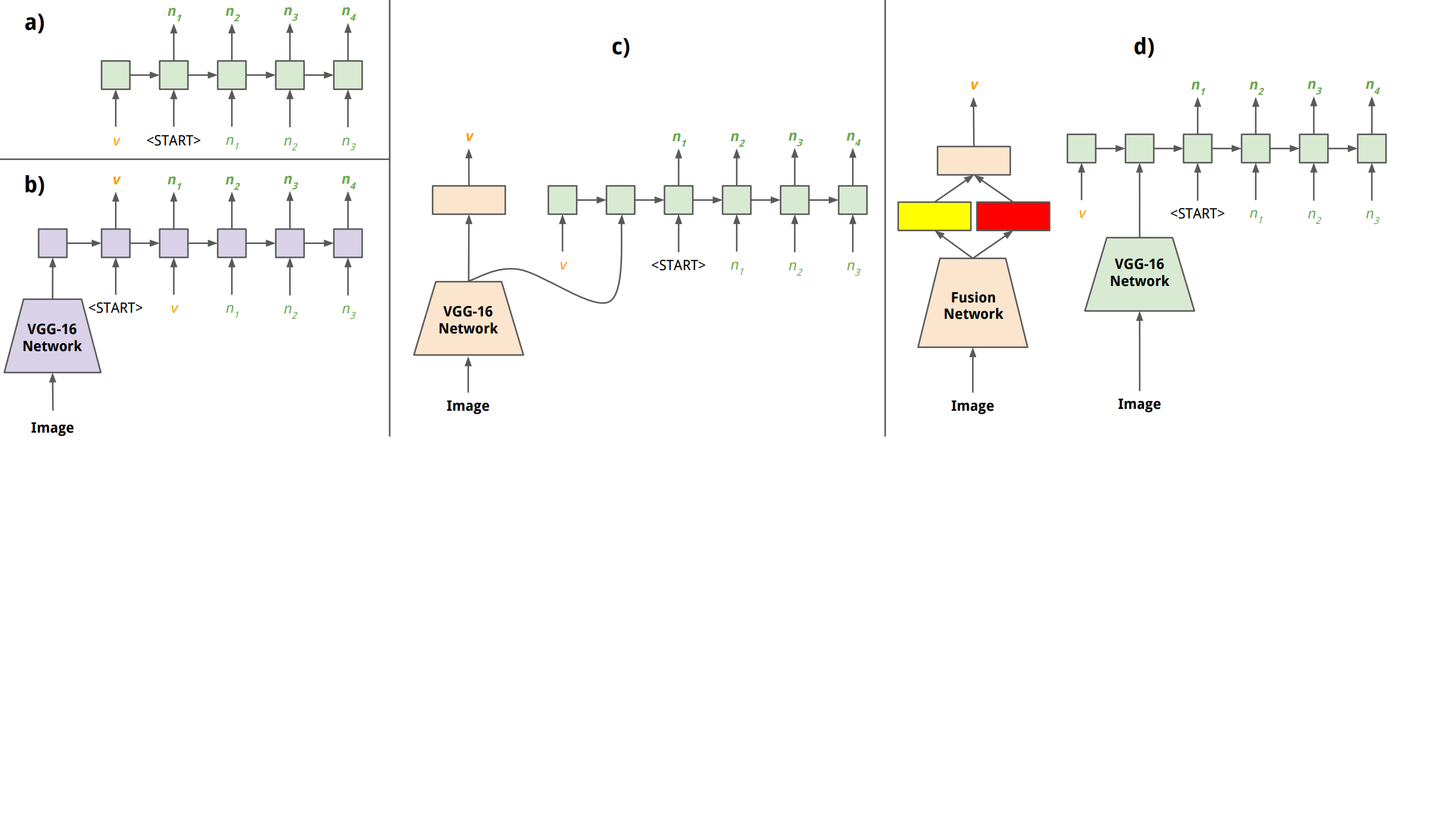}
\caption{The four approaches used for action and noun entity prediction: 
  a) The baseline no-vision model, which only tries to predict noun entities $n_1, \cdots, n_4$ in the chosen arbitrary but fixed semantic role ordering, given the ground truth verb $v$.
  b) Training an RNN which takes image features as input and predicts action, followed by noun entities, 
  c) Training a VGG-16 network for action prediction, and feeding its features 
  to the RNN that predicts nouns associated with the semantic roles, and 
  d) Using separate networks for action and noun entity prediction. 
  Bold colored text (orange and green) indicates training targets.}
  \label{fig:networks}
\end{figure*}

Situations are based on a discrete set of action verbs $V$, noun entities $N$, and semantic roles $R$. Each verb $v\in V$ is paired with a unique frame $f\in F$ derived from FrameNet~\cite{fillmore2003background}, a lexicon for  semantic role labeling. A frame is a collection of semantic roles $R_v\subset R$ which are associated with the verb $v$. For example, the semantic roles \{\emph{Agent}, \emph{Item}, \emph{Tool}, \emph{Place}$\}\subset R$ are associated with the verb \emph{arranging}. In an instantiation of an action in an image, each semantic role is fulfilled by some noun $n\in N\cup \{\varnothing\}$, where $\varnothing$ indicates that the value is either not
known or does not apply. The set of nouns $N$ is derived from WordNet~\cite{miller1995wordnet}. An instance of an action $v$ in an image $I$ forms a realized frame $F_{(I,v)}$ in which each semantic role is associated with some noun $n$, \ie $F_{(I, v)} = \{(r_i, n_i): r_i\in R_v, n_i \in N\cup \{\varnothing\},i=1,\cdots, |R_v|\}$. Finally, a situation $S$ is the pair of action and realized frame for that action, $S=\{v, F_{(I, v)}\}$. The task of situation prediction is to predict an action verb and its associated realized frame given an image. 
Though each image is annotated with a single verb, multiple situations might be applicable for an image due to the choice of nouns used to form a realized frame. For example, one might use the term \emph{countertop} instead of \emph{kitchen} as the noun associated with the semantic role of \emph{Place} in Fig.~\ref{fig:overview}. To account for this multiplicity, the imSitu dataset provides three independently labeled situations per image.

The authors who introduced situtation prediction also proposed a CRF-based approach for the task~\cite{yatskar2016}. They decompose the structured prediction of a situation, $S=\{v, F_{(I, v)}\}$, over the verb $v$ and semantic role value
pairs $(r, n)$ in the realized frame $F_{(I, v)}$. They learn a potential function $\psi_v(v;\theta)$ for every verb, and a potential function for every verb, semantic role, noun entity tuple $\psi_r(v, r, n;\theta)$ ($v\in V$, $r\in R_v$, $n \in N\cup \{\varnothing\}$), where $\theta$ denotes the parameters of the deep neural network used to predict these potentials. The probability of a particular situation $S$ given input image $I$ can thus be represented by:
\begin{equation}
p(S|I; \theta)= \frac{1}{Z} \hspace{1mm}\cdot\hspace{1mm} \psi_v(v|I;\theta) \hspace{1mm} \cdot \hspace{-7mm} \prod_{\substack{(r_i, n_i)\\ r_i\in R_v, n_i \in N\cup \{\varnothing\}}} \hspace{-7mm} \psi_r(v, r_i, n_i|I;\theta).
\label{eq:crf}
\end{equation}
The CRF normalization constant $Z$ required for computing the loss during training is obtained by predicting the potentials for all valid tuples found in the training set and then summing them. The potentials are predicted using a fully connected layer on top of the \emph{fc7} layer of the VGG-16 network~\cite{simonyan14VGG}. During inference time, all valid tuples are scored and ranked. A difficulty with this approach is the large number of potentials that need to be predicted: 504 for all possible verbs and 121,381 for all valid verb, semantic role, noun entity tuples. Further, this model does not explicitly account for the fact 
that nouns are shared across semantic roles, though it is possible that the deep neural network implicitly learns such representations. In order to explicitly enforce the sharing of information and reduce the number of parameters, the follow-up work by Yatskar et al.~\cite{yatskar_semsparsity} further decomposes the potentials as a tensor product over verbs, semantic roles, and noun entities. This makes for a complex model, details of which can be found in~\cite{yatskar_semsparsity}.

We take an alternate view of situation prediction by observing that given a verb $v$, the set of semantic roles $R_v$ associated with it is fixed. For example, given the verb \emph{arranging}, we know that we have to predict relevant noun entities for the semantic roles of $R_{arranging}$=\{\emph{Agent}, \emph{Item}, \emph{Tool}, \emph{Place}\} (see Fig.~\ref{fig:overview}). Conditioned on a given verb, if we assume some arbitrary but fixed ordering over these semantic roles, we can reduce the problem to that of sequential prediction of noun entities corresponding to the semantic roles. 
We decompose $p(S|I; \theta)$ as:
\begin{align}
p(S|I; \theta)&= p\left(v, (r_1, n_1), \cdots, (r_{|R_v|}, n_{|R_v|})|I;\theta\right) \label{eq:seq_pairs}\\
&= p\left(v, n_1, \cdots, n_{|R_v|}|I;\theta\right) \label{eq:seq_order}\\
&= p(v|I;\theta) \hspace{-0.5mm}\prod_{t=1}^{|R_v|} \hspace{-0.5mm} p\left(n_t|v, n_1, \cdots, n_{t-1}, I;\theta\right). \label{eq:seq}
\end{align}

Note that if an arbitrary but fixed ordering is chosen for semantic roles belonging to every verb, then Eq.~\eqref{eq:seq_order} follows from Eq.~\eqref{eq:seq_pairs} as the correspondence of nouns to roles is implicit. In our implementation, we use the semantic role ordering provided in the dataset, which was derived from FrameNet~\cite{fillmore2003background}. 
We explore the sensitivity of methods to the  specific ordering in the experiments of Section \ref{sec:results_imsitu}, and find that the accuracy is affected only to a very small degree. 

We represent each $p\left(n_t|v, n_1, \cdots, n_{t-1}, I;\theta\right)$ in Eq. \eqref{eq:seq} with a softmax over all the noun entities in the training dataset, referred to as the noun vocabulary. This is a standard formulation first introduced for natural language translation~\cite{sutskever2014sequence} and widely adopted for image captioning~\cite{lin2014microsoft,vinyals2015show,vinyals2016show,xu2015show}. Similar to these works, we use a softmax classification loss with the corresponding ground truth noun entity as the target at every prediction step.

It is worth pointing out that both formulations, those of CRF-based structured prediction (Eq.~\eqref{eq:crf}) and sequential prediction (Eq.~\eqref{eq:seq}), are equally powerful in their representational abilities as both model the joint probability of the verb and noun entities in a proposed situation. At inference time, in the CRF approach of~\cite{yatskar_semsparsity,yatskar2016}, all valid tuples of verb and noun entities are evaluated and the most likely one is reported, while in our sequential approach, we perform approximate inference by selecting the most likely noun entity at each step. Despite this limitation, we obtain satisfactory empirical results (we also experimented with beam search but did not see an improvement).


Next, we present the progression of models we developed, starting with a language-only baseline and ending in our highest-performing method illustrated in Figure \ref{fig:overview}.

\smallskip
\noindent{\bf A) No vision, RNN for Nouns.} In order to verify that sequential situation prediction can actually work and that an RNN can memorize the specific ordering of semantic roles for each verb, we propose a basic language-only model that only tries to predict noun entities given the ground truth verb. This model also acts as a strong baseline by exploiting bias in the dataset labeling as it does not use any visual feature input. This model is depicted in Fig.~\ref{fig:networks}a. The ground truth verb is fed in at the first time step. {\em Note that it is essential to feed in the verb at the first time step as the ordering and number of semantic roles for which noun entities are produced is decided by the choice of verb}. 
At the following time step, the RNN tries to predict the noun entity associated with the first semantic role in the arbitrarily selected but fixed ordering, and so on, until a noun entity is predicted for each semantic role for that verb. In line with prior work~\cite{sutskever2014sequence,vinyals2015show}, we feed in the initial verb and the output of the previous time step as a one-hot vector through a word embedding layer. As will be discussed in the next section, this RNN can indeed memorize the arbitrary semantic role ordering to make noun entity predictions in the appropriate order. 

\smallskip
\noindent{\bf B) Shared network, RNN for Actions \& Nouns.} 
The next natural step is to extend the above no-vision model to use image features and predict the action as well. This model is shown in Fig.~\ref{fig:networks}b. After consuming the \emph{fc7} image features from a VGG-16 network at the first time step, the model predicts the action at the second time step and then continues on to predict noun entities. The noun vocabulary (space of all noun entities) is extended with that of possible actions to allow the prediction of both. Note that we use the ground truth action as input during training and the predicted action during testing. At inference time, we enforce that only an action can be predicted at the second time step, followed by noun entities only thereafter. 

\smallskip
\noindent{\bf C) Shared network,\ Actions classifier, RNN for Nouns.} 
Since situation recognition has such a strong up-front dependence on the action verb, the next question we want to explore is whether we can improve performance by breaking off the action prediction into a specialized task, instead of treating it the same as the other roles. It also helps that imSitu has many fewer verbs (504) than noun entities (11K), giving us enough data to train a dedicated action classifier. Accordingly, our second model predicts actions using a separate fully-connected classification layer on top of the \emph{fc7} layer of the VGG-16 network as shown in Fig.~\ref{fig:networks}c. At the first step of the RNN, we feed in the one-hot representation of the action 
(at training time, we use the ground truth action and at test time, the predicted action). At the second time step, we feed in the \emph{fc7} image features to the RNN to predict noun entities. 
Our experiments will investigate how to train the VGG network to get the highest accuracy for the overall task. One option is to train it solely for action prediction and another is to jointly train it for both action and noun prediction. Interestingly, our results in Section~\ref{sec:results_imsitu} will show that the former strategy works better.

\smallskip
\noindent{\bf D) Separate networks,\ Actions classifier, RNN for Nouns.} The lack of success of joint training leads to the question of whether we can do even better by not sharing parameters between action and noun entity prediction.
Accordingly, our final model decouples the two tasks and uses two separate networks that are independently fine-tuned, as depicted in Fig.~\ref{fig:networks}d. For predicting actions, we use the feature fusion network of~\cite{mallya2016learning} which obtained state-of-the-art performance on the HICO dataset~\cite{chao2015hico}. This network (called Fusion in the following) combines local features from detected human boxes and global features from the whole image to make predictions that are then pooled. It defaults to the full image in case no human is detected in the image. As a large number of images in the imSitu dataset feature humans, this is a reasonable choice of architecture. Along with a vanilla RNN for predicting noun entities, we will also report experiments with an attention model based on~\cite{xu2015show} which consumes image features through a soft attention module at each time step. Note that instead of the \emph{fc7} features, the attention-based RNN uses the \emph{conv5} feature map.



\section{Situation Prediction Experiments}
\label{sec:results_imsitu}


\noindent{\bf Implementation Details.} We use the simplified Long-Short Term Memory (LSTM) cell~\cite{hochreiter1997long, zaremba2014recurrent} as our RNN model. We use a single-layer LSTM and with input and hidden layer sizes of 512. We did not observe any significant improvement by using larger layer sizes or more layers. The imSitu dataset has a total of 504 actions and 11,790 noun entities, leading to an LSTM output layer size of 11,790 in the case of models A, C, and D and 11,790+504 in the case of model B. We train all our RNNs with Adam~\cite{kingma2014adam} using an initial learning rate of 4e-4, decayed by a factor of 10 every 28,800 iterations using a batch size of 64. For noun entity prediction, we first train the RNN for 60k iterations. We then turn on fine-tuning for the CNN with an initial learning rate of 1e-5 and use Adam with the same learning rate decay scheme for an additional 100k iterations. The Fusion network~\cite{mallya2016learning} is trained using stochastic gradient descent with momentum using a learning rate of 5e-5 for 70k iterations. Person boxes are detected using the Faster-RCNN~\cite{ren2015faster} with a confidence threshold of 0.8. Similar to~\cite{mallya2016learning}, we use a weighted loss during action prediction, unless otherwise specified. The weight for a class is inversely proportional to its frequency in the training set. Using weighted loss or beam search for noun entity prediction did not help. We only train on the imSitu train set of 75k images. During training, we evaluate the model on the dev set of 25k images and retain the best-performing model. Finally, we evaluate the best model on the imSitu test set of 25k images. All hyperparameters are tuned on the dev set.

\smallskip
\noindent{\bf Metrics.} We evaluate performance on action verb predictions (verb), and (semantic role: noun entity) pair predictions (value, value-all) as well as the average across all measures (mean), as proposed in~\cite{yatskar_semsparsity}. Value-all measures the percentage of predictions for which all of the (semantic role: noun entity) pairs of an action verb matched with at least 1 of the 3 ground truth (GT) annotations, while Value measures the percentage of pairs which matched at least one of the three GT annotations.
We report accuracy at top-1, top-5 action verb predictions and given the GT verb. Similar to~\cite{yatskar_semsparsity}, we also
report performance on examples with ten or fewer samples in the imSitu training set (rare setting).

\begin{table*}[h!]
\setlength{\tabcolsep}{3pt}
\centering
\resizebox{\textwidth}{!}{%
\begin{tabular}{|l|l|c|c|c||c|c|c||c|c||c|}
\hline
\multirow{2}{*}{} & \multirow{2}{*}{} & \multicolumn{3}{|c||}{top-1 predicted verb} & \multicolumn{3}{|c||}{top-5 predicted verbs} & \multicolumn{2}{|c||}{ground truth verbs} & \multirow{2}{*}{mean}\\
\cline{3-10}
& & verb & value & value-all & verb & value & value-all & value & value-all & \\\hline

\multirow{5}{*}{{\bf I)}} & Discrete Classifier~\cite{yatskar2016} & 26.4 & 4.0 & 0.4 & 51.1 & 7.8 & 0.6 & 14.4 & 0.9 & 13.2 \\\cline{2-10}
& Image Regression CRF~\cite{yatskar2016} & 32.25 & 24.56 & 14.28 & 58.64 & 42.68 & 22.75 & 65.90 & 29.50 & 36.32 \\\cline{2-10}
& Tensor Composition CRF~\cite{yatskar_semsparsity} & 31.73 & 24.04 & 13.73 & 58.06 & 42.64 & 22.70 & 68.73 & 32.14 & 36.72 \\\cline{2-10}
& Tensor Comp.\ + Image Reg.\ CRF~\cite{yatskar_semsparsity} & 32.91 & 25.39 & 14.87 & 59.92 & 44.50 & 24.04 & 69.39 & 33.17 &  38.02 \\\cline{2-10}
& Above + Extra 5M Images~\cite{yatskar_semsparsity} & 34.20 & 26.56 & 15.61 & 62.21 & 46.72 & 25.66 & {\bf 70.80} & 34.82 & 39.57 \\\hline\hline

{\bf II)} & \multicolumn{10}{|c|}{\bf Baseline RNN Method} \\\cline{2-11}
{\small Fig.~\ref{fig:networks}a} & No Vision, RNN for Nouns & - & - & - & - & - & - & 52.12 & 17.62 & - \\\cline{2-10}\hline\hline

{\bf III)} & \multicolumn{10}{l|}{\bf \hspace{1cm} Joint Prediction \hspace{0.5cm} -- \hspace{0.5cm} VGG jointly fine-tuned for Action and Noun Prediction} \\\cline{2-11}
{\small Fig.~\ref{fig:networks}b} & VGG, RNN for Actions \& Nouns & 26.52 & 20.08 & 11.80 & 52.37 & 38.32 & 20.90 & 68.27 & 32.67 & 33.87 \\\hline

\multirow{6}{*}{{\Centerstack[l]{ {\bf IV)} \\ {\small Fig.~\ref{fig:networks}c}}}} & VGG, Actions class., RNN for Nouns & 23.04 & 17.65 & 10.70 & 44.63 & 33.18 & 18.83 & 68.98 & 33.73 & 31.34 \\\cline{2-11}
& \multicolumn{10}{l|}{\bf \hspace{1cm} Joint Prediction \hspace{0.5cm} -- \hspace{0.5cm}  VGG fine-tuned for Action Prediction Only} \\\cline{2-11}
& VGG, Actions class., RNN for Nouns & 35.35 & 26.80 & 15.77 & 61.42 & 44.84 & 24.31 & 68.44 & 32.98 & 38.74 \\\cline{2-11}
& VGG, Actions class., RNN for Nouns (reversed) & 35.35 & 26.82 & 15.60 & 61.42 & 44.92 & 24.25 & 68.56 & 32.84 & 38.72 \\\cline{2-11}
& \multicolumn{10}{l|}{\bf \hspace{1cm} Joint Prediction \hspace{0.5cm} -- \hspace{0.5cm} VGG fine-tuned for Action Prediction first, then jointly with Noun Prediction} \\\cline{2-11}
& VGG, Actions class., RNN for Nouns & 34.76 & 26.29 & 15.46 & 60.31 & 44.31 & 24.30 & 68.82 & 33.42 & 38.46 \\\hline\hline

\multirow{10}{*}{{\bf V)}} & \multicolumn{10}{|c|}{\bf Action Prediction Only} \\\cline{2-11}
& VGG, Actions class. (no weighted loss) & 34.43 & - & - & 61.06 & - & - & - & - & - \\\cline{2-11}
& VGG, Actions class. & 35.35 & - & - & 61.42 & - & - & - & - & - \\\cline{2-11}
& Fusion (no weighted loss) & 35.53 & - & - & 63.04 & - & - & - & - & - \\\cline{2-11}
& Fusion & {\bf 36.11} & - & - & {\bf 63.11} & - & - & - & - & - \\\cline{2-11}
& \multicolumn{10}{|c|}{\bf Noun Prediction Only} \\\cline{2-11}
& VGG+RNN for Nouns & - & - & - & - & - & - & 68.57 & 33.12 & - \\\cline{2-11}
& VGG+RNN for Nouns, VGG fine-tuned (ft) & - & - & - & - & - & - & 70.48 & {\bf 35.56} & - \\\cline{2-11}
& VGG+RNN with Attention for Nouns & - & - & - & - & - & - & 69.31 & 33.67 & - \\\cline{2-11}
& VGG+RNN with Attention for Nouns (ft) & - & - & - & - & - & - & 69.87 & 34.69 & - \\\hline
\multirow{3}{*}{{\Centerstack[l]{ {\bf VI)} \\ {\small Fig.~\ref{fig:networks}d}}}} & \multicolumn{10}{|c|}{\bf Separate Action and Noun Prediction} \\\cline{2-11}
& Fusion for Actions, VGG+RNN for Nouns (ft) & \multirow{2}{*}{\bf 36.11} & \multirow{2}{*}{\bf 27.74} & \multirow{2}{*}{\bf 16.60} & \multirow{2}{*}{\bf 63.11} & \multirow{2}{*}{\bf 47.09} & \multirow{2}{*}{\bf 26.48} & \multirow{2}{*}{70.48} & \multirow{2}{*}{\bf 35.56} & \multirow{2}{*}{\bf 40.40} \\
& henceforth ref. to as {\bf Fusion, VGG+RNN} & & & & & & & & & \\\hline
\end{tabular}
}
\caption{Situation prediction results on the full imSitu dev set (see text for detail).}
\label{table:devresults}
\end{table*}

\begin{table*}[h!]
\centering
\small
\begin{tabular}{|l|c|c|c||c|c|c||c|c||c|}
\hline
\multirow{2}{*}{} & \multicolumn{3}{|c||}{top-1 predicted verb} & \multicolumn{3}{|c||}{top-5 predicted verbs} & \multicolumn{2}{|c||}{ground truth verbs} & \multirow{2}{*}{mean}\\
\cline{2-9}
& verb & value & value-all & verb & value & value-all & value & value-all & \\\hline
Image Regression CRF~\cite{yatskar2016} & 32.34 & 24.64 & 14.19 & 58.88 & 42.76 & 22.55 & 65.66 & 28.96 & 36.25 \\\hline
Tensor Comp. + Image Reg. CRF~\cite{yatskar_semsparsity} & 32.96 & 25.32 & 14.57 & 60.12 & 44.64 & 24.00 & 69.20 & 32.97 & 37.97 \\\hline
Above + Extra 5M Images~\cite{yatskar_semsparsity} & 34.12 & 26.45 & 15.51 & 62.59 & {\bf 46.88} & 25.46 & {\bf 70.44} & 34.38 & 39.48 \\\hline\hline
Fusion, VGG+RNN & {\bf 35.90} & {\bf 27.45} & {\bf 16.36} & {\bf 63.08} & {\bf 46.88} & {\bf 26.06} & 70.27 & {\bf 35.25} & {\bf 40.16} \\\hline
\end{tabular}
\caption{Situation prediction results on the full imSitu test set.}
\label{table:testresults}
\end{table*}

\begin{table*}[h!]
\centering
\small
\begin{tabular}{|l|c|c|c||c|c|c||c|c||c|}
\hline
\multirow{2}{*}{} & \multicolumn{3}{|c||}{top-1 predicted verb} & \multicolumn{3}{|c||}{top-5 predicted verbs} & \multicolumn{2}{|c||}{ground truth verbs} & \multirow{2}{*}{mean}\\
\cline{2-9}
& verb & value & value-all & verb & value & value-all & value & value-all & \\\hline
Image Regression CRF~\cite{yatskar2016} & 20.61 & 11.79 & 3.07 & 44.75 & 24.85 & 5.98 & 50.37 & 9.31 & 21.34 \\\hline
Tensor Comp. + Image Reg. CRF~\cite{yatskar_semsparsity} & 19.96 & 11.57 & 2.30 & 44.89 & 25.26 & 4.87 & 53.39 & 10.15 & 21.55 \\\hline
Above + Extra 5M Images~\cite{yatskar_semsparsity} & 20.32 & 11.87 & 2.52 & 47.07 & 27.50 & 6.35 & 55.72 & 12.28 & 22.95 \\\hline\hline
Fusion, VGG+RNN & {\bf 22.07} & {\bf 12.96} & {\bf 3.37} & {\bf 47.83} & {\bf 27.89} & {\bf 6.85} & {\bf 56.38} & {\bf 13.79} & {\bf 23.89} \\\hline
\end{tabular}
\caption{Situation prediction results on the rare portion of the imSitu test set. Along with better verb prediction accuracy, our method also produces more accurate role values given GT verbs, indicating better generalization probably due to the use of shared parameters and word embeddings.
}
\label{table:testresults_rare}
\vspace{-3mm}
\end{table*}

\smallskip
\noindent{\bf Results.}
We report results on the full dev set in Table~\ref{table:devresults}. Section {\bf I} of the table presents results from prior work of Yatskar \etal~\cite{yatskar2016,yatskar_semsparsity}. Their baseline, a method they call the Discrete Classifier, restricts its output space to the 10 most frequent realized frames for each verb. 
The Image Regression CRF uses the formulation of Eq.~\eqref{eq:crf} with an output space of 121,381 for (verb, semantic role, noun entity) tuples + 504 for actions, while Tensor Composition CRF uses a tensor-based potential decomposition in an attempt to reduce the number of parameters. The authors had to combine the potentials produced by both models in order to improve performance, leading to the Tensor Comp.\ + Reg.\ CRF method. Finally, by using five million web-sourced images based on semantic querying~\cite{yatskar_semsparsity} in addition to the 75k train set images, they were able to slightly improve performance.

Our baseline presented in Section {\bf II} of Table~\ref{table:devresults}, corresponding to the architecture of Fig.~\ref{fig:networks}a, shows that RNNs can indeed memorize an arbitrary ordering of semantic roles for each verb and produce relevant noun entities in the correct and corresponding order. Further, by simply exploiting the labeling bias, it beats the Discrete Classifier baseline by a large margin, given the ground truth action verb. 

Section {\bf III} shows results from our next model (Fig.~\ref{fig:networks}b), which tries to predict both the action and noun entities using the same RNN. It improves the value metric by over 16\% given the ground truth verb over our no-vision baseline model, by using information from visual features.

Section {\bf IV} reports the results of separating the action prediction parameters from those of the noun entity predicting RNN (see Fig.~\ref{fig:networks}c). We see a large improvement in action verb prediction accuracy (26.52\% to 35.35\%) as long as we first fine-tune the network for the action task. By simply using features from the network trained for action prediction, we only observe a very small drop in the value metric given ground truth verbs, as compared to jointly fine-tuning for verb and noun entity prediction (68.98\% to 68.44\%). Here, we also try predicting the noun entities in a reversed order so as to determine whether the order affects performance. We clearly see that this has very little effect on accuracy (0.1-0.2\%). However, we cannot rule out that some optimal ordering of semantic roles might exist for every verb. We find that joint fine-tuning, either from the start or later, is detrimental for action verb prediction, leading us to the final models of Sections {\bf V} and {\bf VI}, which use separate networks for action and noun entity prediction.

In Section {\bf V} of Table~\ref{table:devresults}, we compare various methods of separately predicting actions and noun entities. The Fusion network of~\cite{mallya2016learning} outperforms the VGG-16 network at action prediction and using a weighted softmax loss helps in both cases. By using a stand-alone action prediction network, we obtain a top-1 and top-5 accuracy of 36.11\% and 63.11\% in contrast to the previous best of 32.91\% and 59.92\% from~\cite{yatskar2016}, respectively. Even the method from~\cite{yatskar2016} that uses an additional 5 Million images only obtains 34.20\% and 62.21\% accuracies, respectively. 

\begin{figure}[ht!]
  \vspace{-5pt}
  \includegraphics[trim={0cm 6cm 10cm 0cm},clip,scale=0.57]{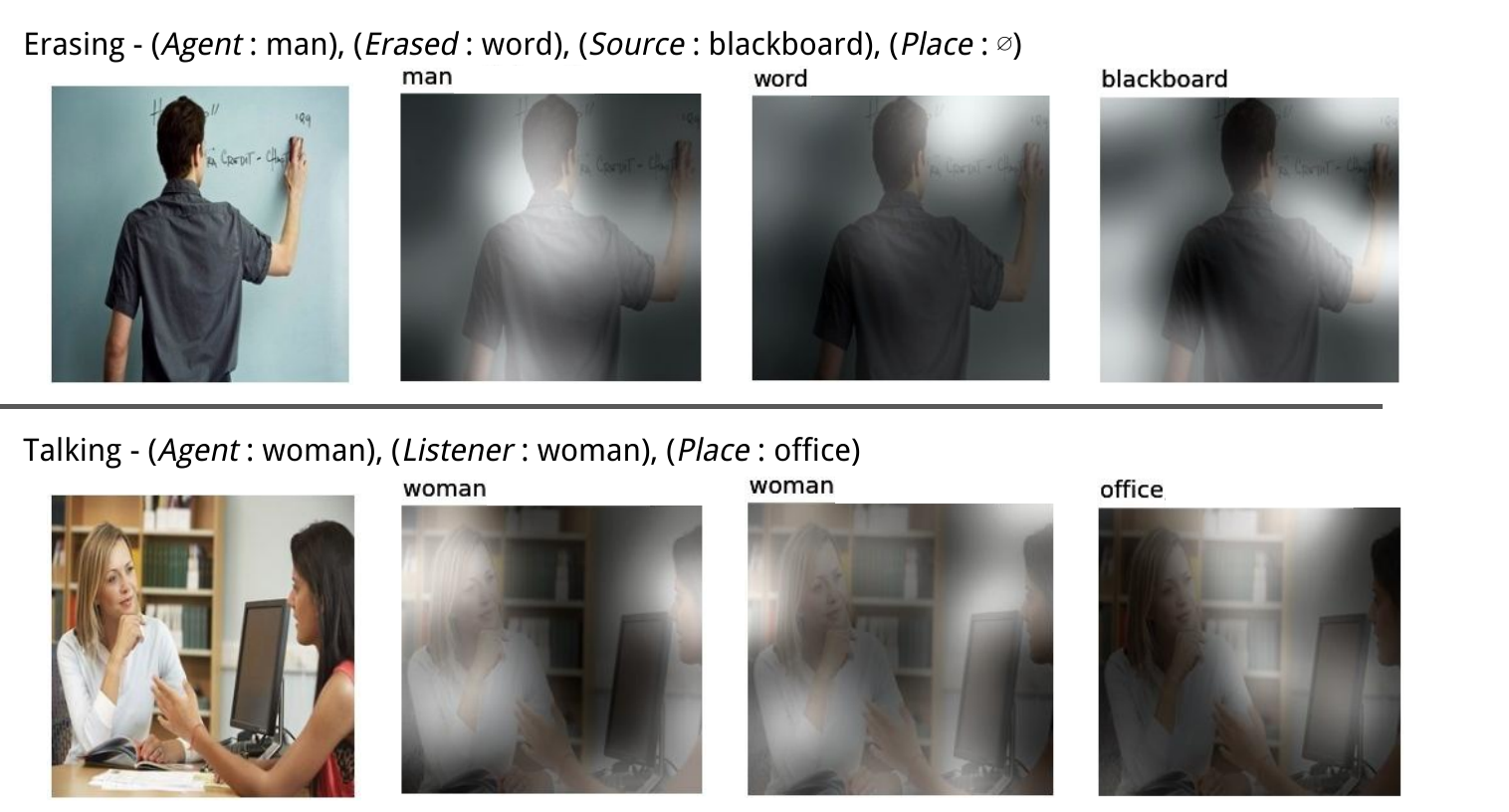}
  \caption{Predicted situations and attention maps associated with produced noun entities. In the top row, attention focuses on the correct regions.
  In the bottom example, attention cannot distinguish between the \emph{Agent} and \emph{Listener} women instances.}
  \label{fig:attention}
  \vspace{6pt}
\end{figure}

Apart from using LSTMs for predicting noun entities, we also try using the soft attention-based architecture of Xu \etal~\cite{xu2015show}. The attention-based RNN works better, as long as we do not fine-tune the underlying VGG-16 network. Turning on fine-tuning makes the simple LSTM architecture work better, in line with results obtained on image captioning~\cite{vinyals2016show}. Figure~\ref{fig:attention} shows some predicted situations and associated attention maps. Qualitatively, attention produces plausible results in simple cases, but is unable to make fine distinctions, e.g., between multiple instances of a noun entity in different roles (bottom row of the figure). 

Finally, we combine our best action prediction and our best noun entity prediction networks to propose our final method referred to as {\bf Fusion, VGG+RNN} (Fig.~\ref{fig:networks}d) in Section {\bf VI} of Table~\ref{table:devresults}. We beat the previous state-of-the-art method trained on the imSitu train set on every metric. Additionally, we also beat the method trained on the extra 5M images, except on the value given ground truth verb metric, on which we lag by just 0.32\%.


Table~\ref{table:testresults} compares our best-performing method against the previous work on the full imSitu test set. We observe a trend similar to that on the imSitu dev test. We improve upon both the top-1 and top-5 verb prediction accuracies by around 3\% and by 1\% (value) and 2.3\% (value-all) on noun entity prediction given ground truth verbs, for methods trained on the imSitu train set. 

Most interestingly, Table~\ref{table:testresults_rare} shows that we also do well on the rare portion of the imSitu test set. We improve upon the top-1 and top-5 verb prediction accuracies by around 2\% and by 3\% respectively. We improve by 3\% (value) and 3.5\% (value-all) on noun entity prediction given ground truth verbs, for methods trained on the imSitu train set. We believe that embedding nouns in a common continuous space during input to RNNs helps to overcome the lack of data and aids in generalization more effectively than the `semantic augmentation' with additional data in the previous method~\cite{yatskar_semsparsity}.

\begin{figure*}[ht!]
  \includegraphics[trim={0cm 0cm 0.75cm 0cm},clip,width=\textwidth]{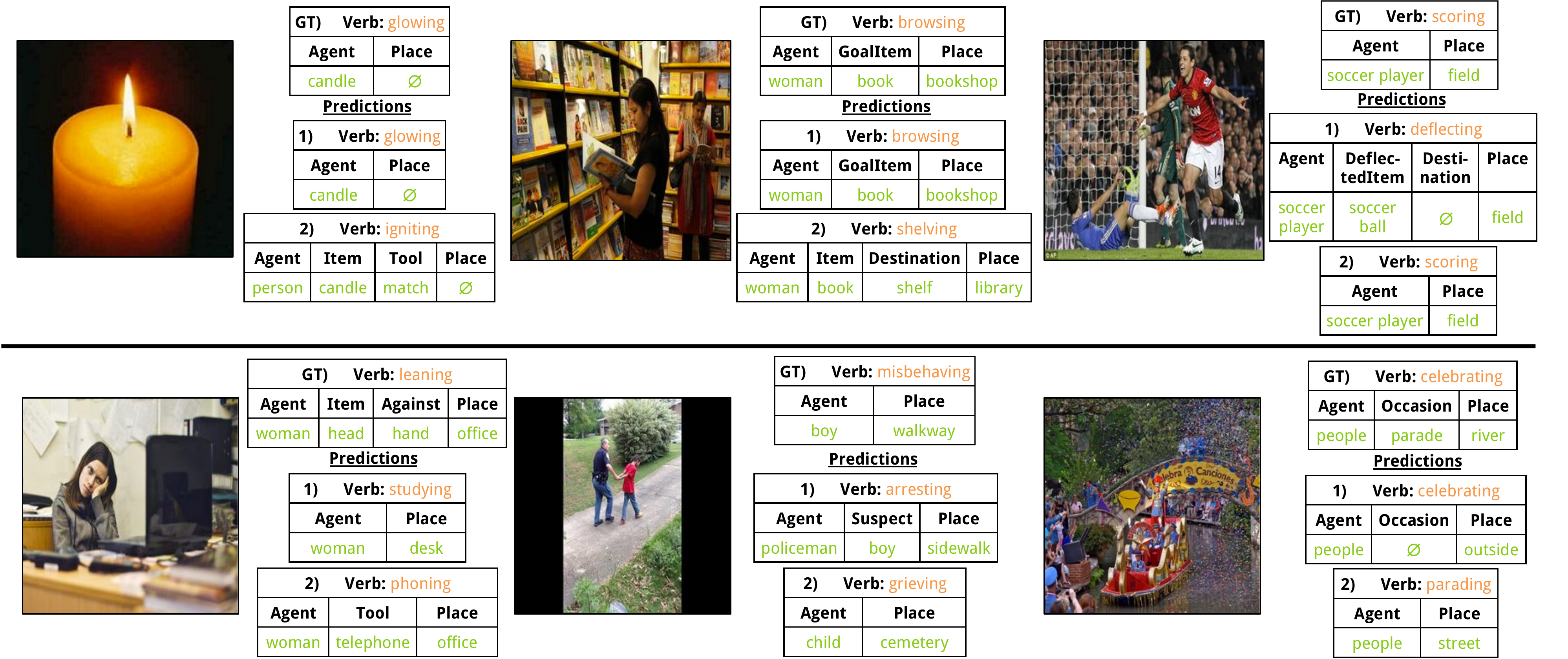}
  \caption{Correct (top row) and wrong (bottom row) predictions on the imSitu test set. One of the three groundtruth labels ({\bf GT}) is shown to the top right of each image. The top 2 predictions (as numbered) are shown below the ground truth. Mistakes can be due to incorrect action verb prediction (bottom row first two images) or incorrect noun entity prediction (bottom right image).}
  \label{fig:imsitu_preds}
\end{figure*}

Finally, Figure \ref{fig:imsitu_preds} shows some correctly and incorrectly predicted situations on the imSitu test set by our best-performing method. While most of the mistakes are due to incorrect action predictions, we observe that mistakes are often reasonable, \eg, \lq arresting\rq\ instead of \lq misbehaving\rq\ in the bottom row, middle image. By analyzing the verb prediction results, we find that we obtain the worst performance on \emph{bothering}, \emph{intermingling}, and \emph{imitating}, which are very contextual and semantic in nature, while those with a clear visual nature such as \emph{erupting}, \emph{shearing}, and \emph{taxiing} obtain high accuracies. The worst noun prediction performance is obtained in cases where multiple nouns can fulfill semantic roles, such as \emph{distributing}, \emph{prying}, \emph{repairing}; while \emph{ballooning}, \emph{taxiing}, \emph{scoring} obtain high accuracies.

\section{Application to Image Captioning}
\label{sec:coco}

One of the key motivations of proposing the task of image situation prediction was to better understand and learn the semantic content of images, beyond mere action recognition~\cite{yatskar2016}. A more structured and nuanced understanding of image semantics is expected to help high-level reasoning tasks such as image captioning and Visual Question Answering (VQA)~\cite{VQA}. In this work, we try to leverage our new state-of-the-art models for action verb and noun entity recognition to improve image captioning performance on the MSCOCO dataset~\cite{lin2014microsoft}.
\begin{figure}[ht!]
  \centering
  \includegraphics[trim={0cm 10.3cm 18cm 0cm},clip,width=0.4\textwidth]{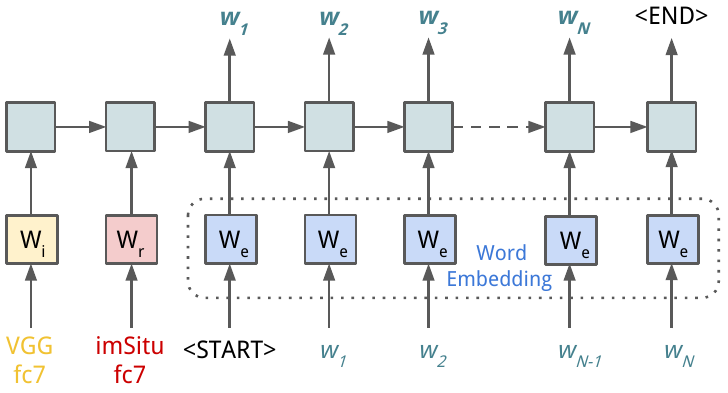}
  \caption{The modified NeuralTalk2~\cite{NeuralTalk2} recurrent neural network that accepts the fc7 feature vector from the networks trained on the imSitu situation prediction task at time step 2. All units with the same color share weights. Bold words $\mathbf{w_1, \cdots, w_N}$ are targets at training time.}
  \label{fig:nt2}
\end{figure}

We modify an off-the-shelf image captioning model, NeuralTalk2~\cite{NeuralTalk2}, by providing it features from our networks as an additional input, as shown in Figure~\ref{fig:nt2}. The vanilla NeuralTalk2 network takes in \emph{fc7} features from a VGG-16 network as input to an RNN through an image embedding layer $W_i$. It then proceeds to output words of the caption one by one till the $<$END$>$ token is predicted or a maximum length (typically 16) is reached. We feed in features from networks trained on imSitu at the second time step, similar to the method proposed in~\cite{yao2016boosting}. We try two types of features: \emph{fc7} features from the VGG-16 network used for noun entity prediction (green network in Fig.~\ref{fig:networks}) and \emph{fc7} features from the VGG-16 network trained for action verb prediction (VGG, fc for Actions of Section {\bf IV} of Table~\ref{table:devresults}). We use features from the VGG-16 network for action prediction instead of the better performing Fusion network because the former produces features from the whole image, while the latter produces features for each detected person box.

\begin{table}[h!]
  \vspace{6mm}
  \centering
  \setlength{\tabcolsep}{2pt}
  \resizebox{\columnwidth}{!}{%
  \begin{tabular}{l|c|c|c|c|c|c|c}
    \hline
    {\bf Methods} & {\bf B@1}& {\bf B@2}& {\bf B@3}& {\bf B@4} & {\bf M} & {\bf C} & {\bf S} \\\hline
    LRCN~\cite{donahue2014long} & 62.8 & 44.2 & 30.4 & 21.0 & - & - & - \\
    img-gLSTM~\cite{jia2015guiding} & 64.7 & 45.9 & 31.1 & 21.4 & 20.4 & 67.7 & - \\
    NIC~\cite{vinyals2015show}$^{\text{\textdagger},\Sigma}$ & 66.6 & 46.1 & 32.9 & 24.6 & - & - & - \\
    img-gLSTM~\cite{jia2015guiding} & 67.0 & 49.1 & 35.8 & 26.4 & 22.7 & 81.3 & - \\
    Hard-Attention~\cite{xu2015show} & {\bf 71.8} & 50.4 & 35.7 & 25.0 & 23.0 & - & - \\
    Soft-Attention~\cite{xu2015show} & 70.7 & 49.2 & 34.4 & 24.3 & 23.9 & - & - \\
    ATT-FCN~\cite{you2016image}$^\Sigma$ & 70.9 & 53.7 & 40.2 & 30.4 & 24.3 & - & - \\\hline
    NeuralTalk2~\cite{NeuralTalk2} (Ours) & 70.8 & 53.7 & 40.1 & 30.1 & 24.5 & 93.0 & 17.3\\
    Image + Actions (Ours) & 71.5 & 54.6 & 40.9 & 30.9 & 24.7 & 94.5 & 17.6 \\
    Image + Nouns (Ours) & 71.5 & {\bf 54.6} & {\bf 41.1} & {\bf 31.1} & {\bf 24.8} & {\bf 95.2} & {\bf 17.7}\\
  \end{tabular}
  }
  \vspace{1pt}
  \caption{\footnotesize Caption generation model performance on the COCO test set (5000 images) of Karpathy \etal~\cite{karpathy2015deep}. {\bf B@N}, {\bf M}, {\bf C}, and {\bf S} indicate  BLEU@N~\cite{papineni2002bleu}, METEOR~\cite{lavie2014meteor}, CIDEr~\cite{vedantam2015cider}, and SPICE~\cite{anderson2016spice} respectively. \textdagger\ indicates a different split of 4000 images and $\Sigma$ indicates an ensemble of models. {\bf Bold} values indicate the highest value for metrics obtained using a single model.}
  \label{table:coco_dev}
\end{table}

\begin{table*}[h!]
  \centering
  \footnotesize
  \begin{tabular}{l|c|c|c|c|c|c|c|c|c|c|c|c|c|c}
    \hline
    \multirow{2}{*}{\bf Methods} & \multicolumn{2}{|c|}{\bf BLEU-1} & \multicolumn{2}{|c|}{\bf BLEU-2} & \multicolumn{2}{|c|}{\bf BLEU-3} & \multicolumn{2}{|c|}{\bf BLEU-4} & \multicolumn{2}{|c|}{\bf METEOR} & \multicolumn{2}{|c|}{\bf ROUGE} & \multicolumn{2}{|c}{\bf CIDEr} \\\cline{2-15}
  & c5 & c40 & c5 & c40 & c5 & c40 & c5 & c40 & c5 & c40 & c5 & c40 & c5 & c40 \\\hline
  ATT-FCN~\cite{you2016image}$^\Sigma$ & 73.1 & 90.0 & 56.5 & 81.5 & 42.4 & 70.9 & 31.6 & 59.9 & 25.0 & 33.5 & 53.5 & 68.2 & 94.3 & 95.8 \\
  OriolVinyals~\cite{vinyals2016show}$^\Sigma$ & 71.3 & 89.5 & 54.2 & 80.2 & 40.7 & 69.4 & 30.9 & 58.7 & 25.4 & 34.6 & 53.0 & 68.2 & 94.3 & 94.6 \\
  MSR\_Captivator~\cite{devlin2015language}$^?$ & 71.5 & 97.0 & 54.3 & 81.9 & 47.0 & 71.0 & 38.0 & 61.0 & 24.8 & 33.9 & 52.6 & 68.0 & 93.1 & 93.7 \\
  Q.Wu~\cite{wu2016value}$^?$ & 72.5 & 89.2 & 55.6 & 80.3 & 41.4 & 69.4 & 30.6 & 58.2 & 24.6 & 32.9 & 52.8 & 67.2 & 91.1 & 92.4 \\\hline
  NeuralTalk2~\cite{NeuralTalk2} (Ours) & 70.6 & 87.9 & 53.2 & 77.8 & 39.2 & 66.1 & 29.0 & 54.7 & 24.2 & 32.4 & 51.9 & 66.0 & 88.1 & 89.1 \\
  Image + Actions (Ours) & 71.1 & 88.6 &  53.9 & 79.0 & 40.1 & 67.7 & 30.1 & 56.7 & 24.4 & 33.0 & 52.3 & 66.8 & 90.1 & 90.7\\
  Image + Roles (Ours) & 71.2 & 88.7 & 54.0 & 79.4 & 40.3 & 68.2 & 30.2 & 57.2 & 24.6 & 33.2 & 52.4 & 67.0 & 90.7 & 91.8 \\
  \end{tabular}
  \caption{Caption generation model performance on the COCO test2014 online leaderboard. We list results that have been published and highlight our implemented baseline and methods. Note that the top methods use ensembles, better model architectures, and other engineering tricks such as scheduled sampling, beyond the scope of this work. The c5 test setting uses 5 reference
captions and c40 uses 40 reference captions. $\Sigma$ indicates an ensemble of models, $?$ indicates unspecified if ensemble.}
  \label{table:coco_test}
\end{table*}

\begin{figure*}[t!]
  \vspace{5pt}
  \centering
  \includegraphics[trim={0cm 0.9cm 2.15cm 0cm},clip,scale=0.7]{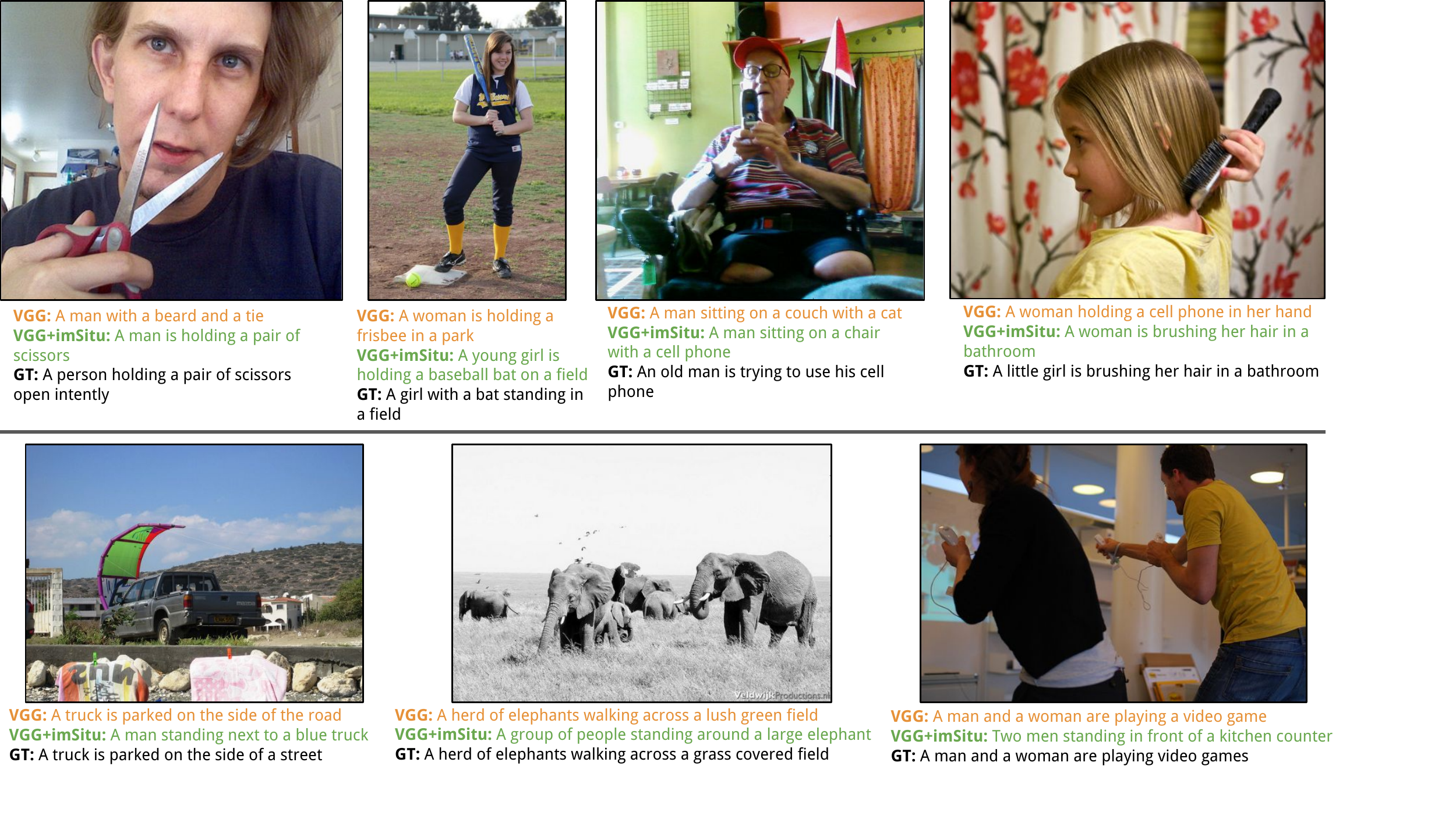}
  \caption{Sample images from COCO test set of Karpathy \etal~\cite{karpathy2015deep} for which adding imSitu features provided the largest gain (top row) and largest drop (bottom row) in CIDEr scores. We also show one of the five ground truth captions that is most similar to the produced captions. We notice that adding imSitu features helps identify and better describe interactions with objects. At the same time, in some of the failure cases, it hallucinates interactions with humans or misidentifies actions.}
  \label{fig:coco_examples}
\end{figure*}

\smallskip
\noindent{\bf Implementation Details and Results.} We use a single-layer LSTM with 512 hidden units and input size of 512. We train our captioning networks on the MSCOCO split of Karpathy \etal~\cite{karpathy2015deep} which has 113,287 training, 5k validation, and 5k test images. We train the RNN and VGG-16 CNN using Adam, with an initial learning rate of 4e-4 and 1e-5 respectively. We train the baseline network in the following recommended stages~\cite{NeuralTalk2,vinyals2016show}: 1) Fine-tune RNN only for 100k iterations, 2) Fine-tune RNN and VGG-16 network for 150k iterations. As shown in Table~\ref{table:coco_dev}, this baseline (NeuralTalk2) obtains a CIDEr score of 93.0 on the test set. We then modify the baseline model to accept an additional imSitu-based feature as input, as shown in Fig.~\ref{fig:nt2} and fine-tune the whole RNN+CNN for another 100k iterations. Beam search of 2 and 3 was found to help the baseline and improved model respectively (recall that it did not help in situation prediction). We see that feeding in imSitu-based features improves the CIDEr score by 2.2 points. Feeding features from the network that produces noun entity predictions (Image+Nouns) works better than features from the action prediction network (Image+Actions). Similar improvements are also observed on the held-out MSCOCO test set as shown in Table~\ref{table:coco_test}. Note that competing methods listed in that table use ensembles and improved architectures to obtain better captioning performance. 

While the quantitative improvements afforded by our additional semantic features are small (and automatic captioning metrics have well-known limitations~\cite{anderson2016spice}), we have qualitatively observed that our captions can describe interactions with objects more accurately, as can be seen from images and captions in the top row of Figure~\ref{fig:coco_examples}. For example, we can correctly identify that a person is holding a baseball bat instead of a frisbee, or a hairbrush instead of a phone. When our model goes wrong (Figure~\ref{fig:coco_examples}, bottom row), it is prone to hallucinating interactions with people.

\section{Conclusion}
\label{sec:conclusion}
This paper framed the recently introduced task of situation recognition as sequential prediction and conducted an extensive evaluation of RNN-based models on the imSitu dataset~\cite{yatskar2016}. Our most important findings are below.
\squishlist
\item RNNs-based methods are a straightforward fit for the task and work quite well.
\item Accurate action prediction is one of the main keys to beating the CRF methods of~\cite{yatskar_semsparsity,yatskar2016}, which do not train an explicit action classifier but predict actions jointly with all the other roles. Further, we found that training a separate action classifier that does not share parameters with noun entity prediction works best. This suggests that the representations needed to predict actions and nouns may be different in non-trivial ways, as it was difficult to fine-tune them jointly.
\item Weakly-supervised attention gives minor improvements but is hard to fine-tune, limiting its absolute accuracy. This is consistent with findings from captioning~\cite{vinyals2016show}. Qualitatively, we found this form of attention to have limited ability to distinguish between entities, indicating the need for advanced attention mechanisms~\cite{kim2017structured}.
\item We have preliminary evidence that situations can help improve captioning quality, though the improvement is currently small. In the future, we will explore better methods to integrate the external knowledge provided by the imSitu dataset into captioning.
\squishend
A limitation of the RNN-based models over CRF-based models is that they cannot produce outputs for verbs unseen at train time as they are unaware of the semantic role ordering associated with the verb. We believe that this can be fixed by making the RNN also output semantic roles, which will be explored in future work.
\smallskip

\noindent {\bf Acknowledgments.} We would like to thank Mark Yatskar for his help with the imSitu dataset. This work was partially supported by the National
Science Foundation under Grants CIF-1302438 and IIS-1563727, Xerox UAC, the Sloan Foundation, and a Google Research Award.

{\small
\bibliographystyle{ieee}
\bibliography{egbib}
}
\end{document}